\pdfoutput=1

\documentclass[11pt]{article}

\usepackage[final]{acl}

\usepackage{times}
\usepackage{latexsym}
\usepackage{CJKutf8}
\newcommand{\cn}[1]{\begin{CJK*}{UTF8}{gbsn}#1\end{CJK*}}

\usepackage[T1]{fontenc}

\usepackage[utf8]{inputenc}

\usepackage{microtype}

\usepackage{inconsolata}

\usepackage{graphicx}

\usepackage{url}
\usepackage{float}          
\usepackage{booktabs}       
\usepackage{tablefootnote}  
\usepackage{multirow}       
\usepackage{pifont}         
\usepackage{amsmath}        
\usepackage{amsfonts}       

\title{\includegraphics[height=0.75\baselineskip]{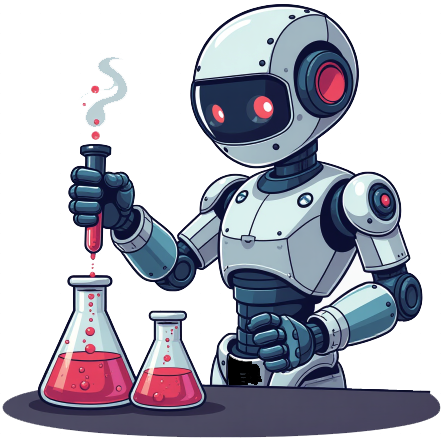} CheMatAgent: Enhancing LLMs for Chemistry and Materials Science through Tree-Search Based Tool Learning}

\author{
  \textbf{Mengsong Wu\thanks{These authors contributed equally to this work.}\textsuperscript{1,2}},
  \textbf{YaFei Wang\footnotemark[1]\textsuperscript{1,3}},
  \textbf{Yidong Ming\textsuperscript{4}},
    \textbf{Yuqi An\textsuperscript{4}},
\\
  \textbf{Yuwei Wan\textsuperscript{4}},
  \textbf{Wenliang Chen\textsuperscript{2}},
  \textbf{Binbin Lin\textsuperscript{3}},
\\
  \textbf{Yuqiang Li\textsuperscript{1}},
  \textbf{Tong Xie\textsuperscript{4}},
  \textbf{Dongzhan Zhou\thanks{Corresponding Author.}\textsuperscript{1}}
\\
\\
  \textsuperscript{1}Shanghai Artificial Intelligence Laboratory,
  \textsuperscript{2}Soochow University, \\
  \textsuperscript{3}Zhejiang University,
  \textsuperscript{4}City University of Hong Kong
\\
  \small{ \texttt{\{wumengsong,wangyafei,zhangdi\}@pjlab.org.cn, \{yidonming2-c,yuqan2-c,yuweiwan2-c\}@my.cityu.edu.hk, }} \\
  \small{ \texttt{wlchen@suda.edu.cn, binbinlin@zju.edu.cn, tong.xie@unsw.edu.au, \{liyuqiang,zhoudongzhan\}@pjlab.org.cn}}
}

\begin{document}
\maketitle

\begin{abstract}

Large language models (LLMs) have recently demonstrated promising capabilities in chemistry tasks while still facing challenges due to outdated pretraining knowledge and the difficulty of incorporating specialized chemical expertise.
To address these issues, we propose an LLM‐based agent that synergistically integrates 137 external chemical tools created ranging from basic information retrieval to complex reaction predictions, and a dataset curation pipeline to generate the dataset ChemToolBench that facilitates both effective tool selection and precise parameter filling during fine-tuning and evaluation.
We introduce a Hierarchical Evolutionary Monte Carlo Tree Search (HE-MCTS) framework, enabling independent optimization of tool planning and execution. 
By leveraging self-generated data, our approach supports step-level fine-tuning (FT) of the policy model and training task-adaptive PRM and ORM that surpass GPT-4o.
Experimental evaluations demonstrate that our approach significantly improves performance in Chemistry QA and discovery tasks, offering a robust solution to integrate specialized tools with LLMs for advanced chemical applications.
All datasets and code are available at \url{https://github.com/AI4Chem/ChemistryAgent}.

\end{abstract}

\section{Introduction}

\begin{figure*}[htb]
  \centering
  \includegraphics[width=0.9\linewidth]{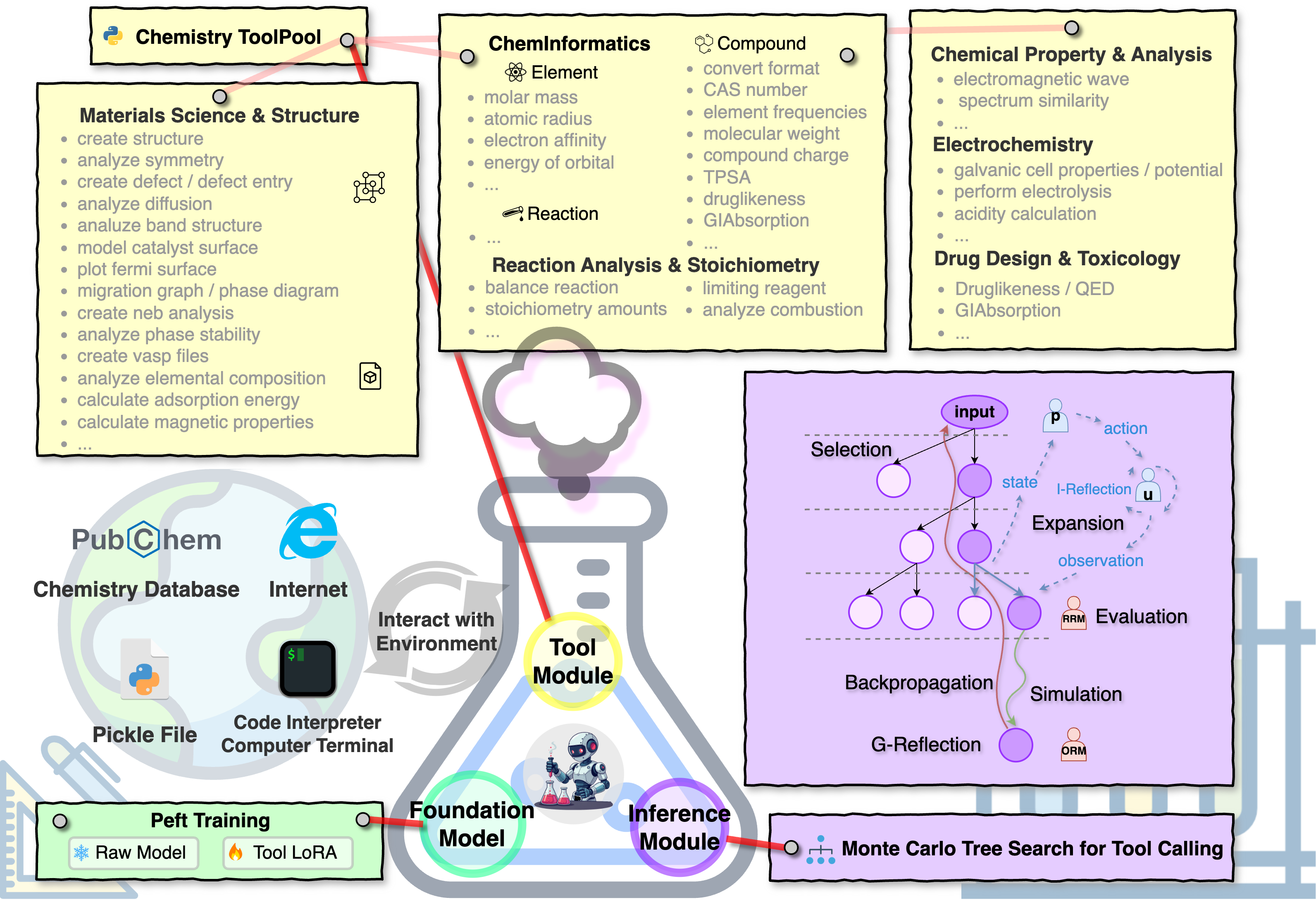}
  \caption{Overview of our CheMatAgent. }
  \label{fig:ChemAgent}
\end{figure*}

\cn{
}



In recent years, Large Language Models (LLMs) have shown considerable promise in tackling chemistry-related tasks~\cite{xue2020x,zhang2024chemllm,mirza2024large}, such as molecule generation and reaction prediction.
However, the expert chemistry knowledge embedded in pretrained models may become outdated and face challenges when applied to real-world scenarios.
One potential solution is the development of LLM-based agents that integrate language models with external, specialized tools to utilize the latest chemistry knowledge. 

Developing LLM-based agents for chemistry has shown significant potential in recent years but there still exists several challenges.
First, existing chemical toolkits rely on specialized cheminformatics software, which is difficult to develop and deploy.
As a result, the number of available tools is limited, which restricts their use in a wider range of chemical tasks.
Additionally, current datasets suffer from poor quality and lack proper evaluation settings.
Even when tools are available, agents struggle with both selecting the right tools and generating accurate parameters due to the specialized knowledge required in chemistry.
These limitations hinder the effectiveness of chemistry-focused LLM agents. 

To address these challenges, we collect a large and diverse set of chemical tools to provide more available tools for LLMs. 
The new toolkit supports a variety of tasks, from simple information queries to complex reaction predictions, which broadens the potential applications of intelligent agents in chemistry.
The code implementation of tools is also in a clear format that is easy to follow, which means more tools can be added to toolpool easily.

A high-quality, diverse meta-dataset ChemToolBench with above tools is then created for fine-tuning the model and serving as the benchmark. 
To construct the comprehensive dataset, we have designed a dataset curation pipeline for self-instruct chemistry Tool Learning data generation.
The dataset includes difficult examples for both tool selection and parameter filling-in, which helps train the model to perform better to call chemistry domain tools.

For better tool calling, 
we introduce an efficient Hierarchical Evolutionary Monte Carlo Tree Search (HE-MCTS) framework. The high-level policy model iteratively explores and refines the tool selection sequence, while the fine-tuned low-level execution model iteratively reflects on execution feedbacks to enhance accuracy. 
Additionally, we leverage self-generated HE-MCTS data alongside the meta-dataset to perform step-level fine-tuning on the policy model, 
and train task-adaptive PRM and ORM as alternatives to GPT-4o. 
Crucially, this training process requires no manual annotation or curation. 
The self-evolving agent, guided by HE-MCTS, autonomously optimizes its performance, demonstrating superior reasoning and execution capabilities.

Our contributions are listed as follows:

\noindent(1) We introduce the largest tool pool in the Chemistry and Materials domain, consisting of 137 tools. An agent augmented with this pool demonstrates superior performance in Chemistry-related QA and discovery tasks.

\noindent(2) We design a dataset curation pipeline tailored for domain-specific tool learning, enabling efficient data generation for fine-tuning. This pipeline supports the construction of the new dataset \textbf{ChemToolBench} for detailed benchmarking.

\noindent(3) We propose \textbf{HE-MCTS}, the Hierarchical Evolutionary Monte Carlo Tree Search framework, that decouples tool planning and execution into separate models. Our framework enables autonomous optimization without manual annotation by leveraging self-generated HE-MCTS data to adopt enhanced step-level FT for the policy model and train the PRM and ORM that surpass GPT-4o in domain-specific task.

\section{CheMatAgent}

Inspired by the success of LLM agents in general scenarios, we attempt to construct an agent for chemistry from scratch.
The foundation LLM of our agent could retrieve and call external tools, and do deep reasoning on complex domain questions.


\subsection{Tools Integration}

This section introduces how to construct executable chemistry toolpools as shown in Figure \ref{fig:ToolIntegration}.
For convenience in agent deployment and evaluation, we hope the tool mainly executes in the local environment and requires slight free online services.
The procedure can be divided into 3 steps as follows.

\subsubsection{Collect Tools from the Internet} \label{sec:tool_collect}
We conduct a survey on former works about chemistry agents / tools\cite{ChemCrow, CACTUS, pymatgen} and also investigate relevant repositories in Github\footnote{\url{https://github.com}}.
Finally we collect tools from 5 sources listed in Table \ref{table:tools_source}: ChemCrow, CACTUS, chemlib, pymatgen, and Chemistry Tools.

\begin{table}[htb]
\centering
\resizebox{0.3\textwidth}{!}{
\begin{tabular}{lc}
\toprule
\multicolumn{1}{c}{\textbf{Source}} & \textbf{Amount} \\ \midrule
ChemCrow \tablefootnote{ChemCrow: \\ \url{https://github.com/ur-whitelab/chemcrow-public} \\ \url{https://github.com/ur-whitelab/chemcrow-runs}} & 8 \\
CACTUS \tablefootnote{CACTUS: \\ \url{https://github.com/pnnl/cactus}} & 10 \\
chemlib \tablefootnote{chemlib: \\ \url{https://github.com/harirakul/chemlib}} & 24 \\
pymatgen \tablefootnote{pymatgen: \\ \url{https://github.com/materialsproject/pymatgen}} & 82 \\
Chemistry Tools \tablefootnote{Chemistry Tools: \\ \url{https://github.com/domdfcoding/chemistry_tools}} & 13 \\
\textbf{In total:} & 137 \\
\bottomrule         
\end{tabular}
}
\caption{Chemistry Domain Tools Source: The number of tools is counted after organization and rewriting in Sections \ref{sec:tool_file_format} and \ref{sec:tool_rewrite_code}.} \label{table:tools_source}
\end{table}

\subsubsection{Organize Tools in Uniform Format} \label{sec:tool_file_format}
To make the tool learning module of the agent extendable, we design a uniform file format for loading python tool packages.
We count all the functions or methods in each package that can be used as tools.
Then we list them in a new JSON file called "tools.json" in each package like in Figure \ref{fig:ToolIntegration}.
The code path for the implementation of the tool is also given in that file.
With the uniform format of each packge, the agent can easily know which tools it has and where to call them.
In the future, more and more chemistry tool packages can be added to our agent without refining the agent framework for compatibility issues as soon as they use the same package organization format as we do. 

\subsubsection{Write Documentation \& Refine Code} \label{sec:tool_rewrite_code}
To make the chemistry toolpool reliable, we also write tool documentation and refine code in the final step of tool integration like shown in Figure \ref{fig:ToolIntegration}.

We write documentation for each chemistry tool so that the agent can better understand the purpose of tools and how to use them.
Besides, chemistry usually contains a variety of compounds, reactions, and other specialized knowledge, with which large models may not be familiar.
So we summarize the input parameters of all the tools with uniform naming.

In Addition, we refine the code implementation of tools to make them easy to use for the agent.
Many tools rely on instances of classes defined in their original Python packages as inputs so it is difficult for the agent to only call the specific tool without declaring other classes.
In order to decouple the tools from their original packages, we adopt two approaches.
\textbf{(1)} For inputs that can be represented with common data types in Python, we convert the original parameters into their corresponding types.
\textbf{(2)} For those can not be easily represented, we read and write them using the pickle file format.

\subsection{Dataset Construction}

A high-quality dataset is the prerequisite for agent fine-tuning and evaluation.
In this section we talk about how to construct the chemistry domain Tool Learning dataset ChemToolBench, trying to design corresponding construction methods by incorporating the characteristics of the chemistry discipline.

\begin{figure*}[htb]
  \centering
  \includegraphics[width=0.7\linewidth]{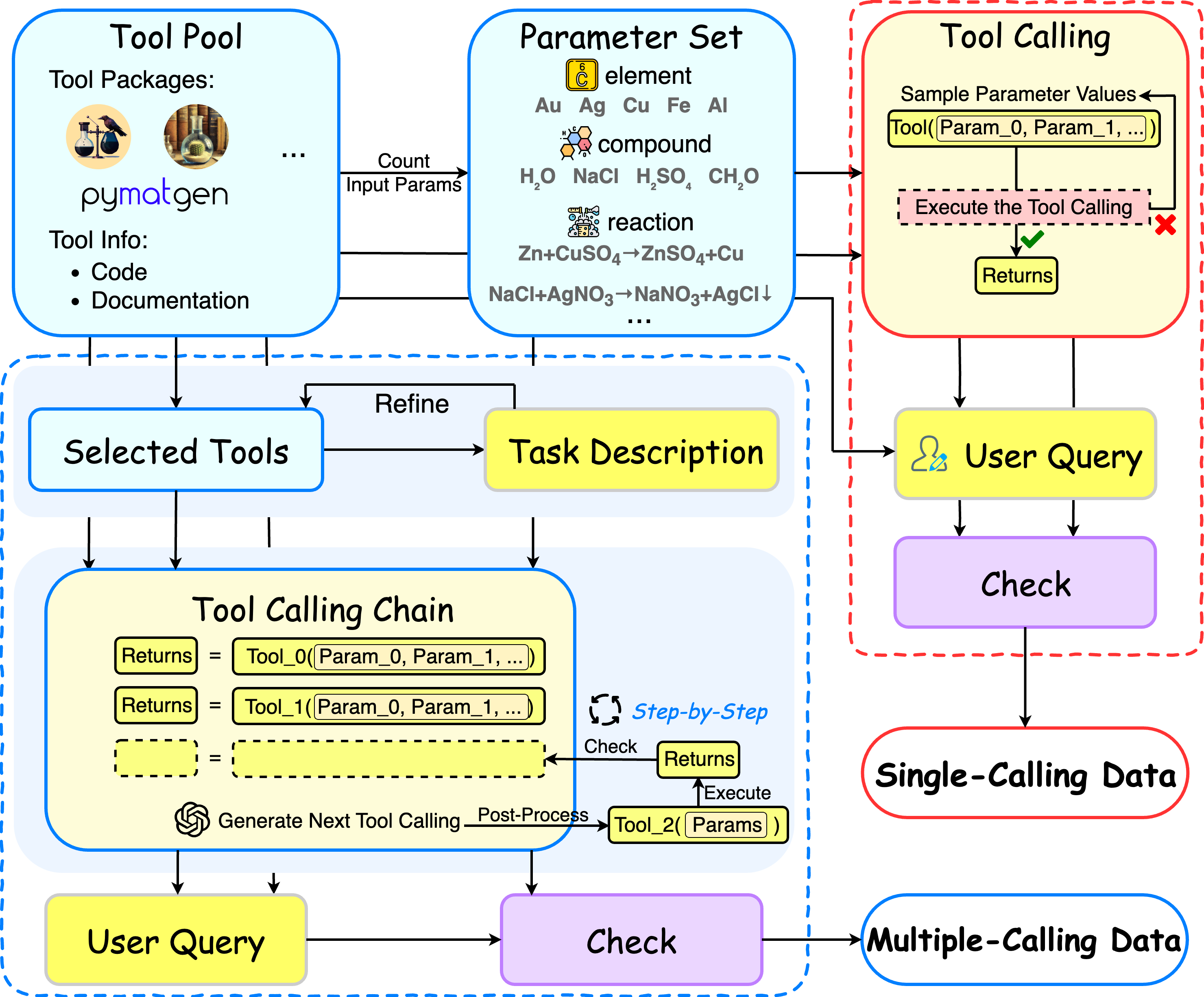}
  \caption{Domain-specific Tool Learning dataset construction pipeline.}
  \label{fig:DatasetConstruction}
\end{figure*}

\subsubsection{Preparation}

In order to better construct data, we generate cases for each kind of parameters.
All parameter names have been standardized in Section \ref{sec:tool_rewrite_code} so that they can be easily categorized.

In our preliminary attempts, we find that LLMs are not good at making up diverse input parameters.
Since the large amount of data constructed by the requirements, the cases returned by the large model over multiple inputs inevitably fall into homogenization.
Besides some parameters involve the user personal privacy, and due to RLHF, the LLMs will simply refuse to return the results, even if they are ordered to generate some virtual examples.

To improve the situation, we find the way to provide some examples of input parameters in prompt like in Figure \ref{fig:DatasetConstruction}.
We use 3 approaches to generate examples for these parameters.
\textbf{(1)} For those chemistry-related concepts, we get examples from the online chemistry database like PubChem\footnote{\url{https://pubchem.ncbi.nlm.nih.gov}}.
\textbf{(2)} For general parameters, we let LLMs to generate as many examples as possible.
\textbf{(3)} For those parameters involving personal privacy like api-key or password, we write code to construct examples.

\subsubsection{Single-Tool-Calling Data}
\label{sec:single_data_generation}

For cases which only need to call single tool, it is relatively easy to generate.
We provide the LLM with the tool and examples of input parameters then the LLM makes up the tool calling as in Figure \ref{fig:DatasetConstruction}.
For tools in packages ChemCrow, CACTUS, chemlib and Chemistry Tools, we try to execute these tool callings to examine the correctness.
For tools in the package pymatgen, we do an exhaustive manual examination after generation.
The same is true for tool calling chains in the next section \ref{sec:multi_data_generation}.
Then we fill in the tool calling and tool information in the prompt to let the LLM generate the user query.
After final manual check, the generation of single-tool-calling data is completed.

\subsubsection{Multiple-Tool-Calling Data}
\label{sec:multi_data_generation}

For cases which need to call multiple tools, we break down the goal into three steps to construct the data.
The quality of the data obtained by letting LLM generate it directly is poor.
The format of the output is often wrong, not to mention the logic of the tool calling chain.
Splitting and subdividing that data generation task as much as possible facilitates better LLMs.

\textbf{STEP 1: Candidate Tool Selection}

The first step is to select several tools from the whole tool pool.
The tool documentation is put into prompt in a disorganized order, and the LLM picks the tools that are relevant from the prompt and generates a rough task description.

\textbf{STEP 2: Tool Calling Chain Generation}

Given candidate tool, input parameter examples and rough task description generated in the last step, the model is asked to generate the tool calling chain step-by-step like in Figure \ref{fig:DatasetConstruction}.

\textbf{STEP 3: User Query Generation}

The third step is to generate the user query according to the tool calling chain and tool documentation.
Finally we do a manual check to examine the correctness and logical soundness.

\subsubsection{Dataset Analysis}
\label{sec:data_analysis}

To the best of our knowledge, we construct the largest and most comprehensive Chemistry Tool Learning dataset.
Our dataset ChemToolBench contains two main splits: \newline
\textbf{Comprehensive Chemistry} split: It has 10441 single-calling data (8353/1044/1044 for train/dev/test) and 2003 multiple-calling data (1623/200/200 for train/dev/test). \newline
\textbf{Materials Science} split: It has 15742 single-calling data (14102/820/820 for train/dev/test) and 1623 multiple-calling data (1187/436 for train/test).

\subsection{The HE-MCTS Framework}
\begin{figure*}[htb]
  \centering
  \includegraphics[width=0.95\linewidth]{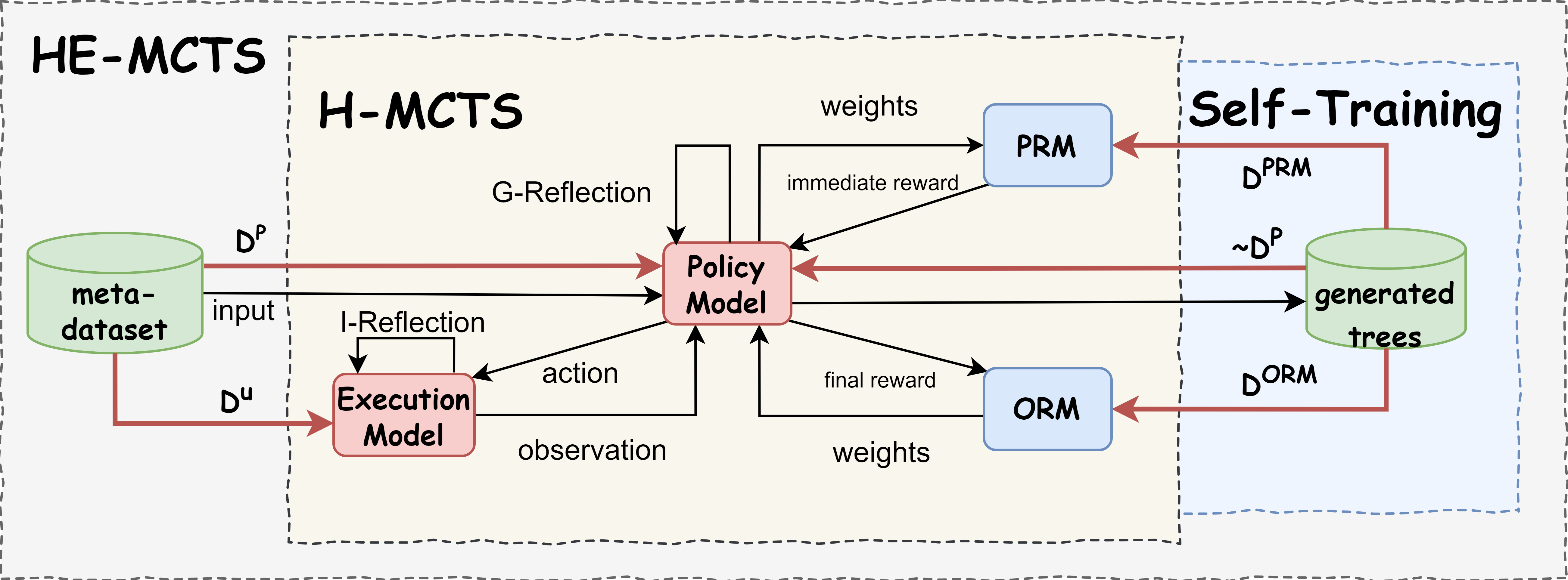}
   \caption{HE-MCTS pipeline.The left part presents the process of Search-Based Hierarchical inferring process. The right part denotes the self-training.}
  \label{fig:HE-MCTS}
\end{figure*}

Our approach, HE-MCTS, is outlined in Figure \ref{fig:HE-MCTS} and developed using four main components.

• Policy Model, which treats tools as aids and integrates tool invocation into a coherent decision process, and Execution Model, which generates specific parameters for each tool invocation, jointly generate step-by-step solutions for each task.

• Hierarchical MCTS, which performs efficiently under the guidance of PRM and ORM.

• Process Reward Model (PRM), which evaluates the quality of any reasoning step, and Outcome Reward Model (ORM), which assesses the quality of the final answer, jointly guide HE-MCTS.

• LLM Self-Training, which leverages HE-MCTS to collect decision trajectories, trains Policy Model on enhanced positive samples, and trains both PRM and ORM on all generated trajectories.  

\subsubsection{Policy Model and Execution Model}
Existing tool agents~\cite{ToolPreference,Toolformer} typically use a single model for both tool planning and execution, though these tasks are inherently different. Tool planning, guided by Tool-Augmented Learning \cite{talm}, requires high-level tasks and tool understanding, while tool execution, guided by Tool-Oriented Learning \cite{tolm}, demands precise operational knowledge. To address this, we decouple tool selection and execution into two components: Policy Model \( p \) and Execution Model \( u \).

At step \( i \), Policy Model generates \( k \) actions $a_i^{j} \sim p(a_i | s^p_{i-1}), \text{for } j = 1, \dots, k$.
The state $s^p_{i-1}$ denotes a partial trajectory $s^p_{i-1} = [x^p, a_1, o_1, \dots, a_{i-1}, o_{i-1}]$. The input \( x^p \) comprises tool selection task prompt, task examples, query \( q \). 
The action $a_i$ comprises thought and tool invocation at step $i$. The valid action space is defined as $A = \{a_i \mid a_i \in T \cup A_n\}$, where $T = \{t_1, t_2, \dots, t_m\}$ denotes the set of available tools, and $A_n$ denotes an aggregated response derived from prefix trajectory. 

Given an action \( a_i^{j} \), the execution result \( o_i^{j} \) is obtained via $o_i^{j} = u(a_i^{j}, s^u_{i-1})$, where the state \( s^u_{i-1} \) is defined as $s^u_{i-1} = [x^u, a_1, o_1, \dots, a_{i-1}, o_{i-1}]$, the input \( x^u \) comprises tool execution task prompt, task examples, query \(q\). 
Execution Model is independently fine-tuned on the dataset \( D^u \), which derived from meta-dataset. Only the log probability of \( \text{parameter\_token} \) is computed.


\subsubsection{Search-Based Hierarchical Reasoning}

In our hierarchical evolutionary framework, we integrate Monte Carlo Tree Search (MCTS) into Policy Model, where each node denotes $s^p_{i-1}$. 

 

\textbf{Uniqueness Enforcement}: Unlike Alphazero\cite{Alphazero}, which promotes diversity through clustering, we enforce uniqueness among sibling nodes by directly filtering out identical execution results, as each result is uniquely determined by the tool and its parameters.

\textbf{Explicit Promotion of Diversity}: Instead of relying on temperature adjustments~\cite{ReST-MCTS*,ETO,SVPO}, we enhance exploration by tracking historical sibling nodes and incorporating diversity prompts. 

\textbf{Prioritization of Unexplored Branches}: Following CPO\cite{CPO}, we prioritize non-terminal nodes to encourage further exploration of unfinished branches during selection. 

\textbf{Adaptive Pruning for Efficient Exploration}: Building on AlphaLLM\cite{AlphaLLM}, we introduce an more adaptive pruning mechanism, which dynamically evaluates nodes using score $I(s^p_i)$ and incorporates Hierarchical Pruning, Soft Pruning, and Fast Recovery to balance search quality and stability. Details are provided in the appendix.

Additionally, we integrate fast-rollout and Global Reflection (Policy-Level), which refines Policy Model by incorporating feedback across multiple search iterations.

Execution Model is directly invoked by Policy Model during the expansion and simulation of the H-MCTS. 
Upon execution failure, Execution Model refines through \textbf{(Tool-Level) Immediate Reflection}, incorporating real-time execution error feedback. 
Once the iterative self-corrective reflection process concludes, the final execution results are returned to Policy Model, which then proceeds with the HE-MCTS evaluation.

\subsubsection{Enhanced Self-Step-FT for Policy Model}
Based on meta-dataset, we construct a step-level dataset for tool selection, denoted as \( D^p = \{(s^p_{i-1}, a_i)\}\). 
Additionally, we construct an enhanced dataset \( \tilde{D}^p = \{(s^p_{i-1}, a_i^{j})\} \) by two strategies.

\begin{figure}[htb]
 \centering
  \includegraphics[width=0.9\linewidth]{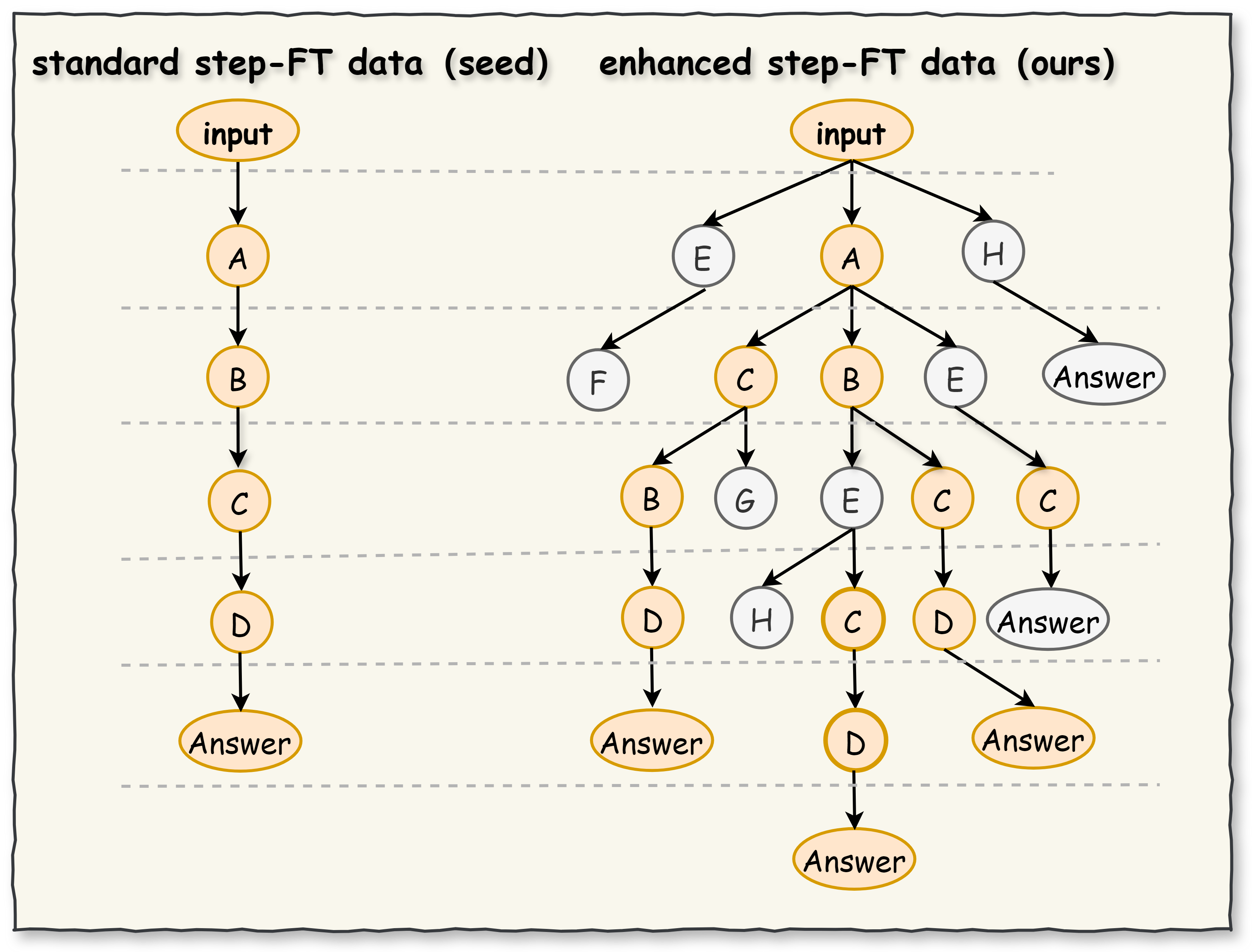}
  \caption{contrast of \(D^p\) and \( \tilde{D}^p\) }
  \label{fig:enhanceddata}
\end{figure}

\textbf{Multi-Path Reasoning and Noise Filtering Strategy}: 
For multi-step tool invocation tasks, LLMs can exhibit multiple valid reasoning paths. 
It is natural to apply a reward-based mechanism that incorporates estimated values to select paths~\cite{SVPO,ReST-MCTS*,AlphaMath} to extract multiple reasoning paths from HE-MCTS trees for fine-tuned. However, such mechanisms do not eliminate noisy actions, potentially leading to errors in credit assignment.
To address this, we filter reasoning paths with meta-dataset, enforcing consistency between each node and the standard invocation chain, ensuring noise-free training labels.
Analysis of search trees reveals that multiplicity stems from the parallel execution of certain tools, with dependencies and interchangeability naturally forming a directed acyclic graph(DAG). Leveraging this structure, an alternative approach is reordering interchangeable tools in meta-dataset, while using GPT to ensure coherent reasoning within each invocation chain.

\textbf{Robustness Reasoning and Noise Retention Strategy}: 
In real-world scenarios, the Policy Model iteratively generates and corrects errors. Discarding paths with incorrect steps or final answers~\cite{CPO,ETO,AlphaLLM} wastes valuable trajectories and weakens its robustness to real-world error patterns.
To address this, we extract nodes from the HE-MCTS tree that follow the correct tool selection strategy, even if their reasoning paths are incomplete or contain errors. Specifically, \( s^p_{i-1} \) may contain incorrect tool invocations, erroneous execution results or perturbed reasoning, while \( a_i^{j} \) remain correct.
Guided by this option, a complementary and more efficient approach perturbs \( D^p \) via rule-based modifications while using GPT to generate corresponding thoughts, observations, or answers.

The comparison between \(D^p\) and \( \tilde{D}^p\) is presented in Figure \ref{fig:enhanceddata}, where each highlighted node can be used to construct a step-FT training sample. 
The loss function for fine-tuning policy\_model is: $\tilde{\mathcal{L}}{p} = \mathbb{E}_{{(s^p_{i-1}, a_i^{j}) \sim \tilde{\mathcal{D}}^p \cup D^p}} \left[ \log p (a_i^{j}|s^p_{i-1}) \right]$.

\subsubsection {Self-Training for PRM and ORM} 
We train two types of self-improving critic models to guide the search process. 
Both PRM and ORM are initialized using Policy Model,  
and their weights remain fixed throughout the HE-MCTS iterations.

\textbf{PRM} The dataset for PRM is constructed as $D^{PRM}=\{(s^p_{i},v_i)\}$
, where $s^p_i$ is sampled from nodes in HD-MCTS trees or the augmented synthetic data. 
$v_i$ is determined on the correctness of \( a_i \) rather than the calibrated value of node $s^p_i$ ~\cite{AlphaMath,ReST-MCTS*,SVPO}. Specifically, if \( a_i \) aligns with the standard tool invocation chain, $v_i$ is 1; otherwise, $v_i$ is 0.
The loss function is: $\mathcal{L}_{PRM}= -\mathbb{E}_{(s^p_{i},v_i) \sim D^{PRM}} \left( V(s^p_i) - v_i \right)^2$.

\textbf{ORM} The dataset for ORM is formulated as $D^{ORM}=\{([q, a_{L}],r_L)\}$, where $q$ and $a_{L}$ originate from the terminal nodes $ s^p_{L}=[q, a_1, o_1, \dots, a_{L-1}, o_{L-1}, a_{L}] (a_{L} \in An) $, sampled from nodes in HD-MCTS trees or the augmented synthetic data. 
$r_L$ is a weighted average of two scores: (1) $r_L^1$, obtained by prompting GPT to assess $a_L$, (2) $r_L^2$, derived from rule-based correctness evaluation of the sequence $[q, a_1, o_1, \dots, a_{L-1}, o_{L-1}]$ using meta-dataset.
The loss function is: $\mathcal{L}_{ORM}= -\mathbb{E}_{([q, a_{L}],r_L) \sim D^{ORM}} \left( R_L - r_L \right)^2$.

\begin{table*}[htb]
\centering
\resizebox{0.9\textwidth}{!}{
\begin{tabular}{lccccccccccc}
\toprule
\multicolumn{1}{c}{\multirow{2}{*}{\textbf{Model}}} & \multirow{2}{*}{\textbf{Format}} & \multicolumn{3}{c}{\textbf{Tool}}  & \multicolumn{3}{c}{\textbf{Param}}  & \multicolumn{3}{c}{\textbf{Return}}  & \multirow{2}{*}{\textbf{Pass Rate}} \\
\multicolumn{1}{c}{}  &  & \textbf{P} & \textbf{R} & \textbf{F1} & \textbf{P} & \textbf{R} & \textbf{F1} & \textbf{P} & \textbf{R} & \textbf{F1} &  \\
\midrule
\textbf{GPT-4o-mini} & 99.83 & 83.22 & 78.86 & 80.98 & 76.04 & 72.19 & 74.06 & 77.05 & 73.13 & 75.04 & 58.00 \\
\textbf{Claude-3.5-S} & 97.42 & 83.64 & 85.37 & 84.50 & 76.03 & 77.53 & 76.77 & 74.96 & 78.54 & 76.71 & 57.00 \\
\textbf{ChemLLM*} & 98.10 & 76.84 & 88.08 & 82.07 & 68.22 & 80.20 & 73.72 & 68.84 & 80.45 & 74.19 & 53.00 \\
\textbf{Qwen-2.5-7B-I} & 51.25 & 64.15 & 41.81 & 50.63 & 56.48 & 37.36 & 44.97 & 29.25 & 37.20 & 32.75 & 22.50 \\
\textbf{Llama-3.1-8B-I} & 75.99 & 59.95 & 38.31 & 46.75 & 52.98 & 33.71 & 41.20 & 40.83 & 34.34 & 37.31 & 15.00 \\
\textbf{Llama-3.1-8B-I*} & 99.20 & 93.10 & 92.21 & 92.65 & 83.85 & 83.15 & 83.50 & 85.03 & 84.90 & 84.96 & 55.00 \\
\midrule

\textbf{GPT-4o-mini-M} & / & 85.06 & 83.57 & 84.31 & 88.47 & / & / & 75.97 & 74.64 & 75.30 & 62.30 \\
\textbf{Claude-3.5-S-M} & / & 89.80 & 86.27 & 88.00 & 86.36 & / & / & 77.55 & 74.51 & 76.00 & 57.06 \\

\textbf{Qwen-2.5-7B-M3} & / & 91.14 & 90.45 & 90.80 & 89.32 & / & / & 81.41 & 80.79 & 81.10 & 67.32 \\

\textbf{Llama-3.1-8B-M0} & / & 75.45 & 78.45 & 76.92 & 85.25 & / & / & 65.09 & 67.68 & 66.36 & 29.19 \\
\textbf{Llama-3.1-8B-M1} & / & 87.74 & 87.32 & 87.53 & 85.92 & / & / & 75.81 & 75.44 & 75.62 & 69.50 \\
\textbf{Llama-3.1-8B-M2} & / & 93.18 & 88.39 & 90.79 & 95.12 & / & / & 88.64 & 84.07 & 86.36 & 72.30 \\
\textbf{Llama-3.1-8B-M3} & / & 93.22 & 91.73 & 92.47 & 92.36 & / & / & 86.09 & 84.72 & 85.41 & 72.20 \\

\bottomrule
\end{tabular}
}
\caption{Main Results on the Multiple-Tool-Calling Comprehensive Chemistry Benchmark. * represents the model fine-tuned with ChemToolBench Comprehensive Chemistry split.}
\label{table:exp_main}
\end{table*}

\section{Experiments}

\subsection{Experimental Setup}

To evaluate the reasoning capabilities of our tool agent in the field of chemistry, we conduct experiments on the ChemToolBench.
Given a user query, the tool retriever would first search relevant tools from the whole tool pool.
Then the LLM judges whether to call these candidate tools.
With the tool calling executed, the LLM takes all return values into consideration and generates the final answer.
For multi-tool calling tasks, we evaluate the performance of the agent under both the Chain-of-Thought (CoT) paradigm and the HE-MCTS paradigm(-M). 
Since the -M agent essentially operates as a multi-agents system, we provide detailed training configurations of each model in Appendix Table \ref{table:exp_dataset_backend}.
For LLMs, we evaluate both commercial models, such as GPT-4o-mini and Claude-3.5-Sonnet, as well as open-source models, including Qwen-2.5, ChemLLM\footnote{\url{https://huggingface.co/AI4Chem/ChemLLM-20B-Chat-DPO}} and the Llama series. 
For tool retriever, we take dense retrievers like all-MiniLM-L6-v2\footnote{\url{https://huggingface.co/sentence-transformers/all-MiniLM-L6-v2}} and NV-Embed v2\footnote{\url{https://huggingface.co/nvidia/NV-Embed-v2}}.

\subsection{Evaluation Metric}

We conduct a comprehensive evaluation of the agent's process reasoning accuracy and result accuracy.
Process reasoning accuracy is assessed in terms of tool selection and tool execution (parameter generation). 
To provide a fine-grained analysis of the agent's reasoning capability, we compute \textbf{Precision, Recall}, and \textbf{F1-score} of tool selection and parameter filling-in. 
Result accuracy is measured using the \textbf{Pass Rate}, which, in the context of question-answering tasks, denotes the proportion of final answers generated by the agent that GPT-4o deems consistent with the reference answers.

\subsection{Results \& Discussion}

\subsubsection{Main Results}
Table \ref{table:exp_main} presents the main experimental results.
For the HE-MCTS, GPT-4o-mini-M and Claude-3.5-S-M exhibit superior than GPT-4o-mini and Claude-3.5-S, primarily due to enhanced tool selection capabilities. 
Our empirical evaluations reveal key advantages of the proposed decoupled hierarchical framework: (1) \textbf{Enhanced Tool Execution Capability}: Compared to end-to-end inference, independently optimizing the Execution Model significantly improves parameter generation accuracy. We attribute this improvement to the substantial reduction in action space enabled by the decoupled framework, allowing the model to focus on specific tasks without unnecessary reasoning over an excessively large search space. (2) \textbf{Positive Impact of Tool Selection on Execution}: As the performance of the Policy Model improves, we observe a minor yet consistent enhancement in the Execution Model. This suggests that more precise tool selection provides a more reliable context for parameter generation, ultimately leading to better execution.

\subsubsection{Ablation Analysis}
For the Policy Model, tool selection capabilities of \textbf{-M0, -M1}, and \textbf{-M2} models exhibit a consistent upward trend, with the \textbf{-M2} models surpassing GPT-4o-mini-M and Claude-3.5-S-M. This result validates the effectiveness of the method we proposed in Section 2.3.3. 

For \textbf{PRM} and \textbf{ORM}, \textbf{-M3} models outperform \textbf{-M2} models, indicating that the PRM and ORM models we trained surpass GPT-4o and Policy Model. 
This advantage is primarily attributed to the improvement of tool execution, as the more specialized critic models reduce redundant sampling in erroneous exploration regions.  

\subsubsection{Generalization Verification}

\begin{table}[htb]
\centering
\resizebox{0.4\textwidth}{!}{
\begin{tabular}{lcllll}
\toprule
\multicolumn{1}{c}{\multirow{2}{*}{\textbf{Model}}} & \multirow{2}{*}{\textbf{Format}} & \multicolumn{1}{c}{\textbf{Tool}} & \multicolumn{3}{c}{\textbf{Param}} \\
\multicolumn{1}{c}{} & & \multicolumn{1}{c}{\textbf{ACC}} & \multicolumn{1}{c}{\textbf{P}} & \multicolumn{1}{c}{\textbf{R}} & \multicolumn{1}{c}{\textbf{F1}} \\
\midrule
\textbf{GPT-4o-mini} & 99.75 & 88.41 & 93.80 & 90.75 & 92.25 \\
\textbf{Qwen-2.5-7B-I} & 54.53 & 43.17 & 85.54 & 44.00 & 58.11 \\
\textbf{Llama-3.1-8B-I} & 98.91 & 75.37 & 88.89 & 81.94 & 85.27 \\
\textbf{Llama-3.1-8B-I*} & 97.93 & 79.51 & 91.32 & 87.50 & 89.37 \\
\bottomrule
\end{tabular}
}
\caption{Single-Calling Results on the Materials Science Benchmark. * represents the model fine-tuned with ChemToolBench Comprehensive Chemistry split.}
\label{table:exp_mat_single}
\end{table}

\begin{table}[htb]
\centering
\resizebox{\linewidth}{!}{
\begin{tabular}{lcllllll}
\toprule
\multicolumn{1}{c}{\multirow{2}{*}{\textbf{Model}}} & \multirow{2}{*}{\textbf{Format}} & \multicolumn{3}{c}{\textbf{Tool}} & \multicolumn{3}{c}{\textbf{Param}} \\
\multicolumn{1}{c}{} & & \multicolumn{1}{c}{\textbf{P}} & \multicolumn{1}{c}{\textbf{R}} & \multicolumn{1}{c}{\textbf{F1}} & \multicolumn{1}{c}{\textbf{P}} & \multicolumn{1}{c}{\textbf{R}} & \multicolumn{1}{c}{\textbf{F1}} \\
\midrule
\textbf{GPT-4o-mini} & 99.90 & 61.37 & 41.31 & 49.38 & 55.91 & 40.21 & 46.78 \\
\textbf{Qwen-2.5-7B-I} & 87.37 & 47.91 & 33.02 & 39.10 & 40.31 & 33.10 & 36.35 \\
\textbf{Llama-3.1-8B-I} & 60.52 & 64.35 & 19.18 & 29.56 & 60.34 & 19.51 & 29.49 \\
\textbf{Llama-3.1-8B-I*} & 94.25 & 76.91 & 73.26 & 75.04 & 71.60 & 65.02 & 68.15 \\
\bottomrule
\end{tabular}
}
\caption{Multiple-Calling Results on the Materials Science Benchmark. * is the same as in Table \ref{table:exp_mat_single}.}
\label{table:exp_mat_multi}
\end{table}

As shown in Table \ref{table:exp_mat_single} and \ref{table:exp_mat_multi}, we also evaluate LLMs on the Materials Science split.
The LLM trained with Comprehensive Chemistry split is also compatible with the other split.
It may suggest that LLMs can learn general chemistry tool knowledge from our dataset ChemToolBench.


\section{Related Works}

The LLM agent with equipped tools has become a hit because it fully extends the application scenarios of the LLMs like science discovery and embodied intelligence.
Tool Learning is one of the important components in an agent.

Several studies focus on constructing tools and datasets for tool learning. 
ToolLLM \cite{ToolLLM} collects APIs from RapidAPI Hub \footnote{\url{https://rapidapi.com/hub}} and employs bottom-up instruction generation, releasing dataset ToolBench.
API-Bank \cite{API-Bank} sets various types of evaluation settings and explores the self-instruct method to construct the dataset.
Seal-Tools \cite{Seal-Tools} tries to generate tools and datasets with LLM from scratch to test the scaling law of tool learning.
ToolACE \cite{ToolACE} introduces a self-evolving API synthesis method and a multi-agent interaction-driven data generation approach, producing 26,507 APIs.
ToolPreference \cite{ToolPreference} trains models using DPO enhances tool usage proficiency.

In general scenarios, many Tool Learning works have gained success in recent years.
Toolformer \cite{Toolformer} demonstrates that LLMs can autonomously learn to use external tools. 
HuggingGPT \cite{HuggingGPT} takes domain-specific language models from Huggingface Hub as tools to solve professional problems.
ToolkenGPT \cite{ToolkenGPT} encodes tools as special tokens in the LLM to decide whether to call a tool during generation.
ToolPlanner \cite{ToolPlanner} simulates real-world user behaviors through multi-granularity instructions and optimizes via path planning.

In scientific scenarios, related explorations are just beginning.
SciAgent~\cite{SciAgent} proposes the scientific reasoning method with domain tools and evaluates it on the new benchmark SciToolBench.
Pymatgen~\cite{pymatgen} builds robust and fast python package for material analysis with many extensions.
ChemCrow~\cite{ChemCrow} integrates 18 expert-designed chemistry tools in the LLM engine to solve tasks like drug analysis and materials design.
It performs better than GPT4 across a range of chemistry tasks while its tools and evaluation questions are limited in amount.
CACTUS\cite{CACTUS} integrates 10 cheminformatics tools to give precise answer in chemistry and molecular discovery questions.

\section{Conclusion}

In this work, we have presented a novel LLM‐based agent specifically tailored for chemical applications by integrating a comprehensive tool pool, an innovative dataset curation pipeline, and an advanced reasoning framework. 
Our approach addresses two major challenges in applying large language models to the chemistry domain: incorporating specialized chemical knowledge and calling multiple tools to solve complex chemistry tasks. 
Furthermore, the introduction of HE-MCTS framework, guided by trained critic models and integrated with an enhanced STEP-FT paradigm, allows our agent to overcome the inherent limitations of the token-by-token decision process in LLMs. 


\section*{Limitations}

Despite the promising results and substantial improvements demonstrated by our approach, several limitations must be acknowledged:

• Computational Overhead: 
Although our HE-MCTS employs various mechanisms to enhance iterative accuracy and efficiency, it inevitably introduces additional computational complexity.
This overhead can hinder real-time applications and may require further optimization to balance efficiency with decision-making accuracy.


• Reliance on Pretrained Knowledge: As with many large language models, our agent effectiveness is partly limited by the potential obsolescence of its pretraining knowledge. Continuous updates and domain-specific fine-tuning are necessary to mitigate this issue and maintain reliability over time.

Addressing these limitations in future research will be crucial for further refining the agent’s performance, ensuring broader applicability, and advancing the integration of specialized tools with large language models in chemical research.


\bibliography{custom}

\appendix
\onecolumn 

\section{More Experimental Details}
\label{sec:apdx_exp}

\subsection{HE-MCTS Model Settings}

\begin{table*}[htb]
\centering
\resizebox{0.9\textwidth}{!}{
\begin{tabular}{lcccc}
\toprule
\textbf{Model} & Policy Model & Execution Model & \textbf{PRM} & \textbf{ORM} \\
\midrule
GPT-4o-mini-M  & /  & /  & /  & gpt-4o  \\
Claude-3.5-S-M  & /  & /  & /  & Claude-3.5-S  \\
Qwen-2.5-7B-M1  & $D^p$  & $D^u$  & $D^p$ & gpt-4o  \\
Qwen-2.5-7B-M2  & $\tilde{D}^p \cup D^p$ & $D^u$  & $\tilde{D}^p \cup D^p$ & gpt-4o  \\
Qwen-2.5-7B-M3  & $\tilde{D}^p \cup D^p$   & $D^u$  & $D^{PRM}$ &  $D^{ORM}$  \\
Llama-3.1-8B-M0  & / & $D^u$ & / & gpt-4o  \\
Llama-3.1-8B-M1  &  $D^p$  & $D^u$  & $D^p$  & gpt-4o  \\
Llama-3.1-8B-M2  & $\tilde{D}^p \cup D^p$ & $D^u$  & $\tilde{D}^p \cup D^p$ & gpt-4o  \\
Llama-3.1-8B-M3  & $\tilde{D}^p \cup D^p$ & $D^u$ & $D^{PRM}$ &  $D^{ORM }$ \\
\bottomrule
\end{tabular}
}
\caption{Training dataset for Different Models}
\label{table:exp_dataset_backend}
\end{table*}

\subsection{Results of Single-Calling Dataset}

\begin{table*}[htb]
\centering
\begin{tabular}{lccccccc}
\toprule
\multicolumn{1}{c}{\multirow{2}{*}{\textbf{Model}}} & \multirow{2}{*}{\textbf{Format}} & \textbf{Tool} & \multicolumn{3}{c}{\textbf{Param}}    & \textbf{Return} & \multirow{2}{*}{\textbf{Pass Rate}} \\
\multicolumn{1}{c}{}   &  & \textbf{ACC}  & \textbf{P} & \textbf{R} & \textbf{F1} & \textbf{ACC}    &  \\
\midrule
\textbf{GPT-4o-mini}  & 100.00   & 93.20   & 96.24   & 94.74   & 95.48   & 90.71   & 86.88  \\
\textbf{Claude-3.5-S} & 98.08  & 92.34   & 97.24   & 95.08   & 96.15   & 90.52   & 80.27  \\
\textbf{ChemLLM-1*}   & 93.97  & 88.31  & 97.06  & 89.02  & 92.86  & 88.12   & 70.98  \\
\textbf{Qwen-2.5-7B-I}  & 64.08   & 56.70  & 91.15  & 61.79  & 73.65   & 54.50   & 49.23  \\
\textbf{Llama-2-7B-C}  & 41.11   & 9.29  & 42.64  & 15.43  & 22.66  & 6.51  & 12.16  \\
\textbf{Llama-3.1-8B-I}  & 98.80  & 83.62  & 93.60   & 87.67  & 90.54  & 81.23  & 63.22  \\
\textbf{Llama-3.1-8B-I*}  & 97.89  & 93.87 & 98.08  & 92.79  & 95.36  & 94.54  & 69.25 \\
\bottomrule
\end{tabular}
\caption{Single-Calling Results}
\end{table*}

\newpage

\subsection{Fine-tuning Results Comparison}

\begin{table*}[htb]
\centering
\begin{tabular}{lccccccc}
\toprule
\multicolumn{1}{c}{\multirow{2}{*}{\textbf{Model}}} & \multirow{2}{*}{\textbf{Format}} & \textbf{Tool}        & \multicolumn{3}{c}{\textbf{Param}} & \textbf{Return} & \multirow{2}{*}{\textbf{Pass Rate}} \\
\multicolumn{1}{c}{} & & \textbf{ACC} & \textbf{P} & \textbf{R} & \textbf{F1} & \textbf{ACC} & \\
\midrule
\textbf{ChemLLM-1*} & & \multicolumn{1}{l}{} & \multicolumn{1}{l}{} & \multicolumn{1}{l}{} & \multicolumn{1}{l}{\textbf{}} & \multicolumn{1}{l}{\textbf{}} & \multicolumn{1}{l}{\textbf{}} \\
\qquad\textbf{- v1} & 93.97 & 88.31 & 97.06 & 89.02 & 92.86 & 88.12 & 70.98 \\
\qquad\textbf{- v2} & 86.21 & 79.89 & 97.49 & 83.69 & 90.07 & 79.98 & 64.37 \\
\qquad\textbf{- v3} & 88.79 & 82.66 & 97.22 & 84.97 & 90.69 & 82.47 & 63.98 \\
\textbf{Llama-3.1-8B-I*} & & & & & & & \textbf{} \\
\qquad\textbf{- v1} & 97.89 & 93.87 & 97.93 & 92.65 & 95.22 & 94.25 & 68.30 \\
\qquad\textbf{- v2} & 97.99 & 93.49 & 97.80 & 92.65 & 95.16 & 94.16 & 72.41 \\
\qquad\textbf{- v3} & 97.89 & 93.87 & 98.08 & 92.79 & 95.36 & 94.54 & 69.25 \\
\bottomrule
\end{tabular}
\caption{Results of LLMs with different fine-tuning model settings on the single-calling benchmark.}
\label{table:exp_NE_single}
\end{table*}
\begin{table*}[htb]
\resizebox{\textwidth}{!}{
\begin{tabular}{lccccccccccc}
\toprule
\multicolumn{1}{c}{\multirow{2}{*}{\textbf{Model}}} & \multirow{2}{*}{\textbf{Format}} & \multicolumn{3}{c}{\textbf{Tool}} & \multicolumn{3}{c}{\textbf{Param}} & \multicolumn{3}{c}{\textbf{Return}} & \multirow{2}{*}{\textbf{Pass Rate}} \\
\multicolumn{1}{c}{} & & \textbf{P} & \textbf{R} & \textbf{F1} & \textbf{P} & \textbf{R} & \textbf{F1} & \textbf{P} & \textbf{R} & \textbf{F1} & \\
\midrule
\textbf{ChemLLM-1*} & & \multicolumn{1}{l}{} & \multicolumn{1}{l}{} & \multicolumn{1}{l}{} & \multicolumn{1}{l}{} & \multicolumn{1}{l}{} & \multicolumn{1}{l}{} & \multicolumn{1}{l}{} & \multicolumn{1}{l}{} & \multicolumn{1}{l}{} & \multicolumn{1}{l}{} \\
\qquad\textbf{- v1} & 98.50 & 56.98 & 89.51 & 69.64 & 50.43 & 81.60 & 62.34 & 51.74 & 82.51 & 63.60 & 46.00 \\
\qquad\textbf{- v2} & 86.46 & 68.19 & 89.98 & 77.59 & 59.42 & 81.04 & 68.57 & 53.85 & 82.19 & 65.07 & 46.50 \\
\qquad\textbf{- v3} & 98.10 & 76.84 & 88.08 & 82.07 & 68.22 & 80.20 & 73.72 & 68.84 & 80.45 & 74.19 & 53.00 \\
\textbf{Llama-3.1-8B-I*} & & & & & & & \textbf{} & & & & \\
\qquad\textbf{- v1} & 99.18 & 60.25 & 93.00 & 73.13 & 51.60 & 83.57 & 63.81 & 55.06 & 85.69 & 67.04 & 57.50 \\
\qquad\textbf{- v2} & 96.81 & 94.56 & 91.26 & 92.88 & 85.03 & 82.16 & 83.57 & 84.21 & 83.94 & 84.08 & 57.00 \\
\qquad\textbf{- v3} & 99.20 & 93.10 & 92.21 & 92.65 & 83.85 & 83.15 & 83.50 & 85.03 & 84.90 & 84.96 & 55.00 \\
\bottomrule
\end{tabular}
}
\caption{Results of LLMs with different fine-tuning model settings on the multiple-calling benchmark.}
\label{table:exp_NE_multi}
\end{table*}

\textbf{v1} means the fine-tuning dataset contains no negative cases.
\textbf{v2} means the fine-tuning dataset contains negative cases for both single and multiple callings.
\textbf{v3} means the fine-tuning dataset contains negative cases for only multiple callings.

\newpage

\section{Algorithm Details of H-MCTS}

\subsection{Process of H-MCTS}
\begin{figure*}[htb]
  \includegraphics[width=\linewidth]{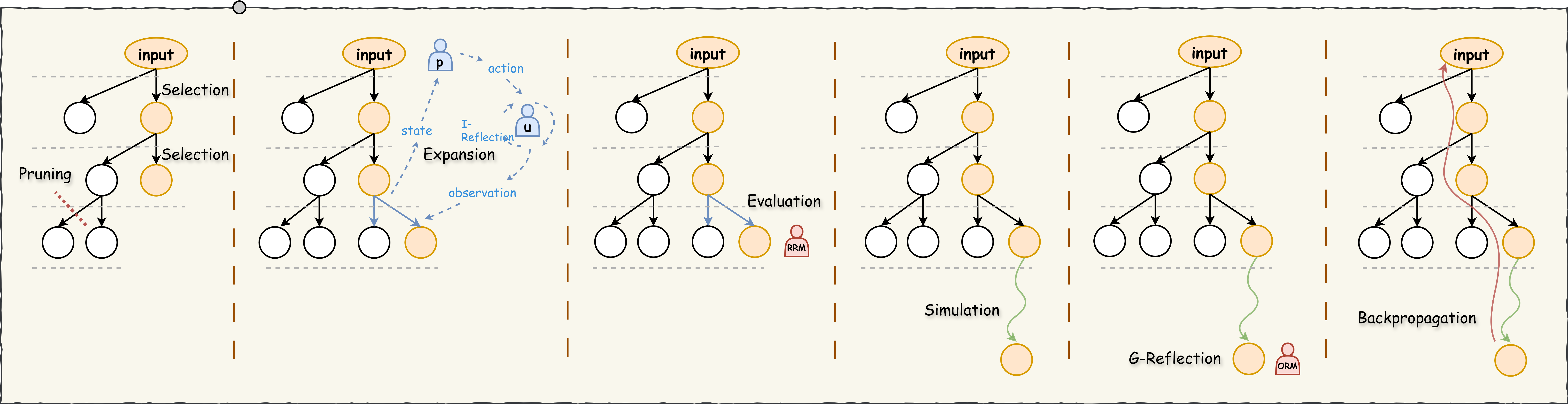}
  \caption{H-MCTS process}
  \label{fig:H-MCTS}
\end{figure*}

Hierarchical Monte Carlo Tree Search (H-MCTS) is a sampling-based search algorithm. It iteratively constructs a search tree by repeating six phases as illustrated in Figure \ref{fig:H-MCTS}: Expansion, Evaluation, Selection, Simulation, Reflection, and Backpropagation.

\textbf{Expansion}: Policy Model generates $k$ child nodes. 
To enhance the efficiency of exploration, $T$ is constrained to retrieved tools. 
To mitigate redundancy, uniqueness is enforced among child nodes, 
ensuring parent node does not generate duplicate children. 
Furthermore, to enhance the distinctions between sibling nodes, leverage diversity prompts.

\textbf{Evaluation}: The PRM initializes a scalar score $V_i$ for newly expanded nodes:
\begin{equation}
V_i = V(s^p_i) = PRM(s^p_i)
\end{equation}
This score is used in the Selection phase to compute the Upper Confidence Bound for Trees (UCT) value of nodes and serves as a reference for choosing starting points in the subsequent Simulation phase.

\textbf{Selection}: It recursively selects nodes from the root based on the Upper Confidence Bound\cite{UCB} (UCB) which allows the search to prioritize high-value nodes while still encouraging the discovery of new solutions:
\begin{equation}
    UCT(s^p_{i-1},a_i^{j}) = V_i^{j} + C \cdot \sqrt{\frac{\ln(N(s^p_{i-1}))}{1+N(s^p_{i-1}, a_i^{j})}}
\end{equation}
where $N(s^p_{i-1})$,$N(s^p_{i-1}, a_i^{j})$ denote visit counts. $V_i^{j}$ is initialized by PRM. The hyperparameter $C$ is an exploration coefficient. 
To promotes exploration of unfinished branches, prioritize non-terminal nodes over terminal ones.
To maintain efficient search space, nodes with low information gain and value are adaptively pruned before selection.

\textbf{Simulation}: The Policy Model predicts subsequent actions from selected leaf node until reaching a terminal node $s^p_{L}$, where $s^p_{L} = [q, a_1, o_1, \dots, a_{L-1}, o_{L-1}, a_{L}](a_{L}\in A_n)$. The reward $R_L$ is assigned by $ORM(q, a_{L})$. 
To expedite trajectory simulation and expansion, a single node is sampled at this stage. 

\textbf{Global Reflection(Policy-Level)}: If agent fails to yield a correct answer, the Policy Model performs failure analysis and generates recommendations to guide subsequent iterations.

\textbf{Backpropagation}: Starting from \( s^p_L \), updates propagate along the path back to \( s^p_0 \):
\begin{equation}
    N(s^p_i) \gets N(s^p_i) + 1
\end{equation}
\begin{equation}
    V(s^p_i) \gets V(s^p_i) + \frac{R_{L} - V(s^p_i)}{N(s^p_i)}
\end{equation}
where final reward $R_{L}$ can source heuristic rule or external reward function, like ORM: 
\begin{equation}
R_{L} = ORM(q,a_{L}), \quad a_{L} \in A_n
\end{equation}

\subsection{Details of Adaptive Pruning mechanism}
\textbf{Scoring Mechanism} The core of pruning is the evaluation of node importance. We compute a comprehensive node score:
\begin{equation}
I(s^p_i) = \alpha V(s^p_i) + \beta U(s^p_i)
\end{equation}
where $V(s^p_i)$ is the value of node $s_i$, measuring the historical search gains, $U(s^p_i)$ is the uncertainty estimation of node $s^p_i$, based on Information Gain \cite{pearl1984heuristics}(IG), which quantifies the node's importance in the overall search strategy.

For a node $s^p_i$ with visit count $N(s^p_i)$ and a child set $C(s^p_i)$, Information Gain (IG) is defined as: 
\begin{equation}
U(s^p_i) = H(s^p_i) - \sum_{c \in C(s_i)} \frac{N(c)}{N(s^p_i)} H(c)
\end{equation}
where $H(s^p_i)$ represents the entropy\cite{silver2016mastering} of node $s^p_i$, computed based on search trajectory, indicating the uncertainty of decision-making at that node. A higher information gain suggests a greater impact on the search strategy.

Pruning is guided by an adaptive threshold $\tau(i)$, such that nodes with scores below the threshold are pruned.

\textbf{Hierarchical Pruning}
The pruning threshold $\tau(i)$ dynamically adjusts based on search depth $i$:

• Shallow search($i < D_{\text{early}}$): A lower pruning threshold encourages broader exploration, reducing premature pruning effects
\begin{equation}
  \tau(i) = \tau_0 \cdot \left( 1 - \lambda \frac{i}{D_{\max}} \right)
\end{equation}

• Deep search($i > D_{\text{early}}$): The pruning threshold increases, prioritizing high-value paths
\begin{equation}
  \tau(i) = \tau_0 \cdot \left( 1 + \lambda \frac{i}{D_{\max}} \right)
\end{equation}

where \( \tau_0 \) is the initial pruning threshold controlling overall pruning intensity, \( \lambda \) is a hyperparameter regulating threshold variation, \( D_{\max} \) is the maximum search depth, ensuring progressive pruning refinement.

\textbf{Soft Pruning}
To mitigate search loss from mispruning, Soft Pruning\cite{gelly2011monte} retains pruned nodes with a certain probability. If \( I(s^p_i) < \tau(i) \), the node is retained with probability:
\begin{equation}
P_{\text{retain}} = e^{-\kappa (\tau(i) - I(s^p_i))}
\end{equation}
where \( \kappa \) controls the pruning probability decay rate, allowing nodes close to the threshold to have a higher retention probability.

\textbf{Fast Recovery}
To prevent excessive pruning from limiting search effectiveness, we introduce a Fast Recovery mechanism\cite{spirtes1991algorithm} :

• Pruned Node Logging: Maintain records of pruned nodes, including Score $I(s^p_i)$, Pruning depth $i$, Visit count $N(s^p_i)$.

• Detect Search Degradation: If the search reward drops significantly compared to the best path:
\begin{equation}
  \frac{V_{\text{best}} - V_{\text{current}}}{V_{\text{best}}} > \epsilon
\end{equation}
where $V_{\text{best}}$ is the average value of the best search path, $V_{\text{current}}$ is the average value of the current search path, $\epsilon$ is the recovery threshold.

Restore recently pruned high-score nodes from history records to reintroduce potentially valuable search directions.



\newpage

\section{Tool Integration Procedure}
\label{sec:apdx_tool}

\begin{figure}[htb]
  \centering
  \includegraphics[width=0.5\linewidth]{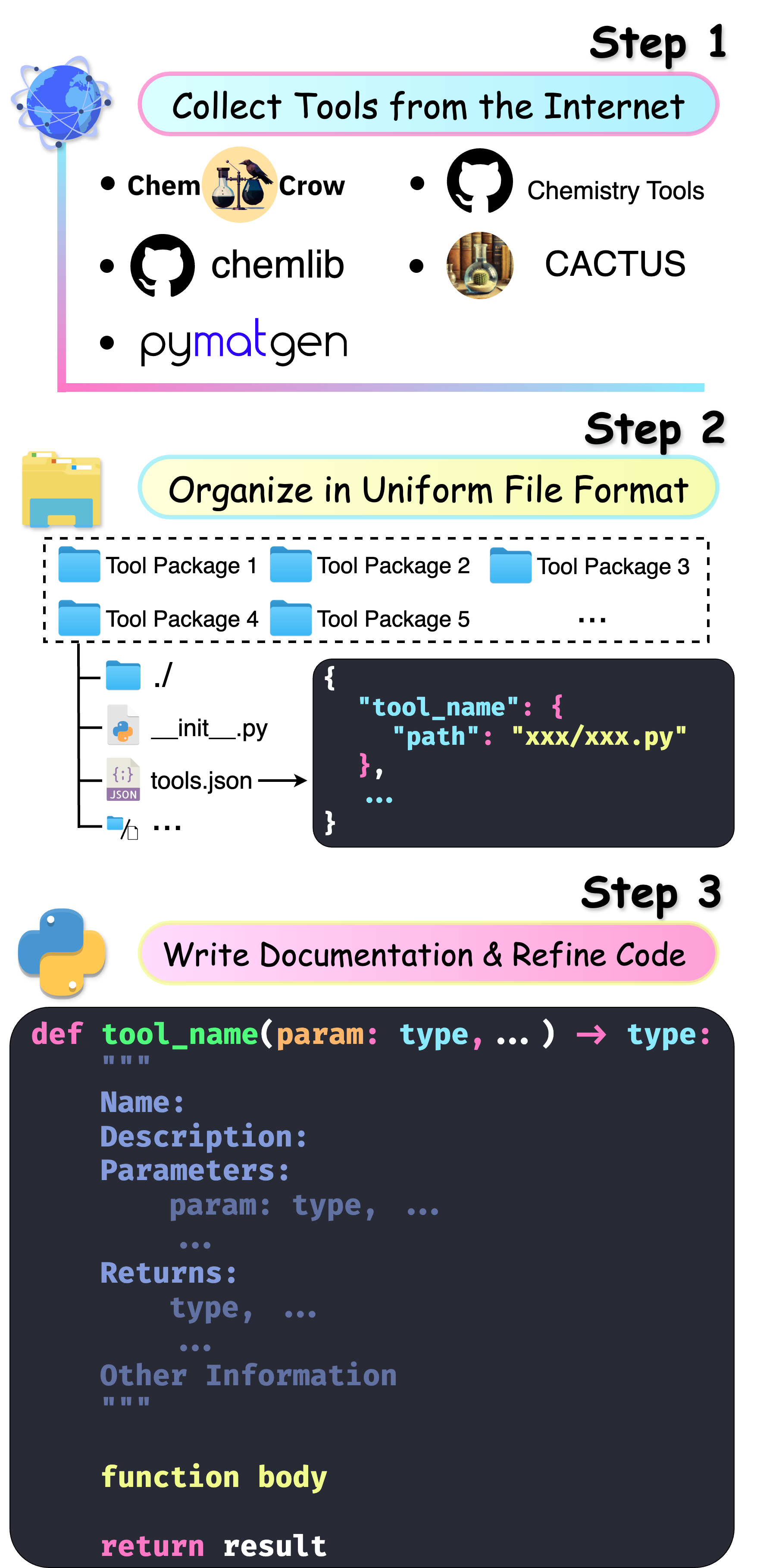}
  \caption{Tool Integration Procedure in 3 steps.}
  \label{fig:ToolIntegration}
\end{figure}

\end{document}